\documentclass{ppai}

\usepackage{ppai}

\usepackage[utf8]{inputenc} 
\usepackage[T1]{fontenc}    
\usepackage{hyperref}       
\usepackage{url}            
\usepackage{booktabs}       
\usepackage{amsfonts}       
\usepackage{nicefrac}       
\usepackage{microtype}      
\usepackage{lipsum}
\usepackage{float}
\usepackage{multirow}
\usepackage{graphicx}
\usepackage{subcaption}
\usepackage{dirtytalk}
\usepackage{makecell}
\graphicspath{ {./figures/} }

\title{Efficient and Private: Memorisation under differentially private parameter-efficient fine-tuning in language models}

\author{
 Olivia Ma \\
  Department of Computing\\
  Imperial College London\\
  \texttt{om323@ic.ac.uk} \\
   \And
 Jonathan Passerat-Palmbach  \\
  Flashbots \& \\ Imperial College London \\
  \texttt{j.passerat-palmbach@imperial.ac.uk} \\
  \And
 Dmitrii Usynin \\
  AI for Medicine \\
  Technical University of Munich\\
  Department of Computing \\
  Imperial College London \\
  \texttt{du216@ic.ac.uk} \\
}

\begin{document}
\maketitle
\begin{abstract}
Fine-tuning large language models (LLMs) for specific tasks introduces privacy risks, as models may inadvertently memorise and leak sensitive training data. While Differential Privacy (DP) offers a solution to mitigate these risks, it introduces significant computational and performance trade-offs, particularly with standard fine-tuning approaches. Previous work has primarily focused on full-parameter updates, which are computationally intensive and may not fully leverage DP’s potential in large models. In this work, we address these shortcomings by investigating Parameter-Efficient Fine-Tuning (PEFT) methods under DP constraints. We show that PEFT methods achieve comparable performance to standard fine-tuning while requiring fewer parameters and significantly reducing privacy leakage. Furthermore, we incorporate a data poisoning experiment involving intentional mislabelling to assess model memorisation and directly measure privacy risks. Our findings indicate that PEFT methods not only provide a promising alternative but also serve as a complementary approach for privacy-preserving, resource-efficient fine-tuning of LLMs.
\end{abstract}

\section{Introduction}
Training of accurate machine learning (ML) models often requires access to large, well-curated datasets of high quality. 
Obtaining access to this data becomes more challenging if the learning task is categorised as \say{high risk} under the recently introduced EU AI act \cite{wei2022emergent, european_commission_2021_ai_act}.
Moreover, this issue has become particularly important due to a recent surge in interest towards large language models (LLMs), which are often trained on large undisclosed datasets (some of which may inadvertently include private information).
This phenomenon has raised a number of data privacy and governance concerns when training such models on sensitive, personally-identifying data \cite{das2024security}.
This is primarily because it was recently demonstrated in a number of prior works that LLMs are capable of unintentionally memorising the data they have previously been trained or fine-tuned on \cite{song2019auditing, Carlini2021extract, mireshghallah2023quantifying, zeng2023exploring}.
One popular method to alleviate these privacy concerns is the use of differentially private (DP) \cite{abadi2016dpsgd, dwork2014algorithmic} model training.
DP is a property of a randomised algorithm, which makes it approximately invariant to the contributions of a single data point, resulting in a provable upper bound on how much information can be memorised for each participant \cite{dwork2014algorithmic}.
This can typically be achieved through the addition of carefully calibrated noise (based on the sensitivity of an algorithm) to the output of a randomised algorithm.
In the context of deep learning, this is manifested in a widespread use of differentially private stochastic gradient descent (DP-SGD) \cite{abadi2016dpsgd}.
DP-SGD relies on the addition of noise to the gradients of the model during training, where the amount of noise required can be calibrated through gradient clipping to obtain a bounded sensitivity.
These two steps, in turn, can significantly lower the utility of the final model \cite{ponomareva2022training, bagdasaryan2019differential}. 

Furthermore, as the gradient norms of samples need to be clipped individually, this creates additional computational costs associated with the use of DP-SGD, resulting in higher memory requirements and longer training times.
This issue is particularly profound for LLMs, given their already large computation requirements \cite{yu2021differentially, bu2022differentially, li2021large}.
One family of methods frequently used to reduce the cost of fine-tuning LLMs on the downstream tasks is Parameter-Efficient Fine-Tuning (PEFT) \cite{han2024parameter, chandra2024parameter}.
These methods aim to reduce the computational overhead of model fine-tuning by updating only a small subset of parameters.

While PEFT methods were primarily developed to improve the efficiency of training, recent research highlights that certain PEFTs may have long-lasting effects on the resulting model's behaviour.
For example, LoRA tends to exhibit lower degrees of forgetting compared to full-model fine-tuning at the cost of reduced memorisation capacity\cite{Biderman2024LoRA}. 
These properties suggest that PEFT methods may reduce task-specific memorisation and associated privacy risks.
These benefits, alongside their ability to address the computational challenges of DP, such as excessive noise addition and prolonged training times, PEFT methods present a promising alternative to full-model fine-tuning \cite{chandra2024parameter}.
However, the impact of PEFTs on the privacy of the trained model remains mostly under-explored, as prior works primarily focused on efficiency without a thorough empirical evaluation of privacy leakage. 

In this work, we aim to address these gaps by empirically evaluating PEFT methods under DP training, specifically focusing on their interplay with DP and how the combination of individual techniques can be used to reduce privacy risks alongside preserving the model utility.
Our main contributions are as follows:
\begin{itemize}
    \item We evaluate three most widely used PEFT methods, namely: Adapters, LoRA, and (IA)\(^3\) under DP training, comparing their task-specific performance and privacy leakage against standard fine-tuning methods on the sentiment analysis and natural language inference tasks.
    \item We empirically evaluate the memorisation capacity of the models under DP by analysing the worst-case inputs to the model under privacy attacks.
    \item Our results demonstrate that PEFT methods can achieve a promising balance between computational efficiency and privacy preservation, showing that these methods can be used alongside DP for better efficiency while still giving provable privacy guarantees.
\end{itemize}

\section{Background and Related work}
Differential Privacy (DP) \cite{firstdp2006} is one of the most widely used information-theoretic frameworks for formal preservation of privacy under public data releases.
This is achieved through the addition of noise to the output of the randomised algorithm, thereby limiting the impact a single individual can have on its output and the amount of information that is released is controlled through the privacy budget defined by $\varepsilon$ (lower - more private).
The DP definition used in this work, known as \((\varepsilon, \delta)\)-DP, extends the initial definition of $\varepsilon$-DP, by using an additional parameter \(\delta\), allowing a small probability of privacy compromise \cite{zhu2017differentially}: 
\[
\Pr[A(D_1) = x] \leq e^{\varepsilon} \Pr[A(D_2) = x] + \delta
\]
where \(D_1\) and \(D_2\) are datasets differing by one record (through an addition/removal of an individual under \textit{unbounded}-DP or a replacement under \textit{bounded}-DP), \(A\) represents a randomized algorithm, and \(\varepsilon\) and \(\delta\) are privacy parameters, with smaller \(\varepsilon\) indicating stronger privacy.
In ML this can be achieved through the addition of randomised noise to the gradients of the trained model (whose norms need to be clipped to a pre-defined threshold of $C$ in order to bound the sensitivity and add the appropriately calibrated amount of noise) under DP-SGD \cite{abadi2016dpsgd}.
However, differentially private training introduces significant computational and performance trade-offs. 
The additional noise and gradient clipping in DP-SGD limits the information the model can learn during training and can impose substantial memory and processing overhead, particularly challenging for large-scale models such as LLMs \cite{yu2021differentially, bu2022differentially, li2021large, yousefpour2021opacus}.

To allow efficient fine-tuning of LLMs, a number of PEFT methods that modify only small parts of the model, such as Adapters \cite{houlsby2019adapter}, LoRA \cite{hu2021lora} and (IA)\(^3\) \cite{liu2022ia3} have previously been proposed.
A more detailed discussion on the attributes of individual PEFTs can be found in \cite{han2024parameter}.
Each PEFT method targets different parts within the transformer architecture: Adapters add (and fine-tune) compact task-specific layers within the feedforward (FF) layers.
LoRA applies low-rank approximations to decompose the weight matrices in transformer models, instead of updating weights directly, it tracks changes to weight matrices during fine-tuning.
(IA)\(^3\) selectively scales key activations with learned vectors, with the vectors injected in the attention and FF layers to focus on important features.
PEFT techniques have demonstrated successful applications across various tasks by significantly reducing the number of trainable parameters while maintaining or enhancing model performance.
For example, Adapter and LoRA achieved state-of-the-art results in speech emotion recognition and graph-based tasks, showcasing their ability to adapt large models efficiently \cite{Gui2023G-Adapter, Feng2023PEFT-SER}.
(IA)\(^3\) excells in few-shot scenarios, outperforming in-context learning while introducing minimal new parameters \cite{liu2022ia3}.
Combined approaches like AdaMix and LoRAPrune rely on a mixture of modular adaptations and pruning techniques to enhance PEFT's efficiency and scalability for complex tasks \cite{Wang2022AdaMix, Zhang2023Pruning}

Several approaches have previously been proposed to alleviate the computational overhead of DP fine-tuning in LLMs.
Yu et al. \cite{yu2021differentially} introduced a meta-framework that incorporates PEFT methods, such as Adapters and LoRA, into DP fine-tuning.
Their work demonstrated improvements in memory and computational efficiency while maintaining performance across a range of NLP tasks. 
However, the authors did not empirically assess the privacy risks associated with these methods, leaving open questions about how PEFT methods interact with DP mechanisms in terms of privacy leakage.
Our work bridges this gap by empirically testing how PEFT methods behave under DP, especially focusing on privacy risks such as memorisation.

DP-forward, proposed by Du et al. \cite{du2023dpforward}, represents another attempt to reduce the costs of DP fine-tuning by modifying the model's embedding matrices using non-i.i.d, noise, instead of perturbing all model parameters.
While this approach achieves reductions in memory usage and training time, it is currently limited to encoder-only models, making implementation adaptation to other architectures challenging. 

Another approach, DP-BiTFiT \cite{bu2022differentially} focuses on fine-tuning only the bias terms of a model, rather than the full set of model parameters.
This method significantly reduces the number of trainable parameters while preserving utility under DP.
However, updating only the bias terms is often suboptimal for highly complex tasks like generative modelling or tasks requiring deep feature transformations.

Membership Inference Attacks (MIAs) have been previously used the means of assessing such privacy leakage in fine-tuned models \cite{murakonda2020ml, song2020introducing}.
Originally introduced by Shokri et al. \cite{shokri2017membership}, MIAs allow adversaries to infer whether a specific data point was included in a model’s training set by analysing its outputs, leveraging the differences in how the model responds to training versus non-training data.
Research has demonstrated that fine-tuned large language models (LLMs) are particularly susceptible to MIAs \cite{fu2023practical, mireshghallah2022empirical, mireshghallah2023quantifying, truex2019effects}.

There exist several variants of MIAs.
Black-box MIAs rely solely on model outputs and include approaches such as loss-based, entropy-based, and confidence-based attacks.
Loss-based MIAs compare loss values between training and test data, entropy-based MIAs measure prediction uncertainty, and confidence-based MIAs assess prediction confidence.
They usually exploit the difference in model behaviour between training data and unseen (test) data.
In contrast, white-box MIAs assume full access to model parameters and gradients, allowing for more precise attacks, though these are less practical in real-world scenarios \cite{usynin2023memorisation, dionysiou2023sok}.
A recent advancement in this field, namely the Robust Membership Inference Attack (RMIA) \cite{zarifzadeh2023low}, relies on access to a population dataset (real data from the same distribution as training data), employing likelihood ratio comparisons instead of training a large number shadow models.
RMIA achieves improved performance, further strengthening MIA's effectiveness as a more resilient metric for evaluating privacy leakage in more realistic adversarial settings.

Complementary to MIAs, poisoning attacks provide another perspective on model vulnerabilities by injecting maliciously crafted samples into the training data to manipulate the behaviour of the trained model.
These attacks can induce incorrect associations or biased patterns, revealing the model's propensity to internalize specific, often harmful, information \cite{chen2017targeted, qiang2024poison, chen2024amplifying}.
In the context of evaluation of model's privacy (and its ability to memorise individual sampled) a related technique, namely the canary insertion \cite{carlini2019secret} can be used.
Canaries are unique, identifiable phrases or tokens inserted into the training data to quantify the model's ability to memorise sensitive information (e.g. credit card numbers), as demonstrated in studies by Carlini et al. \cite{carlini2019secret, Carlini2021extract}.

\section{Assessing Privacy Leakage in PEFT}

In this work, we empirically assess the potential privacy leakage that occurs during fine-tuning of LLMs, and study the interactions between parameter efficient fine-tuning methods behave with DP with respect to both the utility of the model and the privacy of the training data.
To this end, we have conducted a number of \textit{black-box loss-based MIAs} in non-DP and DP settings, which analyses model output losses to infer the membership of a data point \cite{jayaraman2019evaluating}.
The Area Under the Curve (AUC) scores serves as the primary metric for quantifying the privacy leakage \cite{mattern2023membership}.

To investigate the \say{worst-case} memorisation, we created a number of canaries as part of model training, which were made to deliberately be more atypical (and hence more prone to memorisation) \cite{carlini2019secret}.
We achieved this by intentionally mislabelling a set of data points to perform MIAs on these at the end of training \cite{qiang2024poison}.
Specifically, we selected $30$ instances each from the training and test sets, modifying their labels to create a small subset of canary-like data points.
The number of flipped samples was kept small to avoid significantly impacting the overall performance on correctly labelled data, while still allowing us to observe the model's memorisation capacity on these specific data points.
After training under both DP and non-DP settings, we conducted a loss-based MIA on the entire dataset and then focused on the AUC scores of this poisoned subset.
High AUC scores on this subset would indicate that the model retains memorised associations for these mislabelled instances, thereby demonstrating a tendency towards memorisation.

Finally, we investigated whether the reduced memorisation risks observed in PEFT-trained methods could be attributed to their smaller number of trainable parameters alone.
To test this, we increased the parameter counts for Adapter and LoRA and evaluated the effectiveness of MIAs on models fine-tuned on IMDb and QNLI (see Table \ref{tab:param_variation} for details).

\begin{table}[!h]
\centering
\begin{tabular}{|l|l|c|}
    \hline
    \textbf{Method} & \textbf{Variation} & \textbf{Trainable Parameters} \\
    \hline
    \multirow{3}{*}{Adapter} & Bottleneck (32) & 599,424 \\ 
                              & Bottleneck (128) & 2,370,048 \\
                              & Bottleneck (512) & 9,452,544 \\
    \hline
    \multirow{3}{*}{LoRA} & \( r \) (8) & 739,586 \\ 
                           & \( r \) (96) & 2,361,602 \\
                           & \( r \) (480) & 9,439,490 \\
    \hline
\end{tabular}
\caption{\footnotesize Parameter variation configurations for Adapter and LoRA in the fine-tuning experiments. For Adapter, the variations correspond to different bottleneck sizes in the FF layer, while for LoRA, \( r \) denotes the rank of low-rank matrices added to the attention layers. Note that these configurations differ slightly for full-model fine-tuning under DP, where BERT-base and DistilBERT have 109,483,778 and 66,561,794 trainable parameters, respectively, with positional embeddings frozen for compatibility with \cite{yousefpour2021opacus}.}
\label{tab:param_variation}
\end{table}

Our experimental setup primarily utilized pre-trained LLMs, specifically \textit{DistilBERT} \cite{sanh2019distilbert}, with \textit{BERT-base} \cite{devlin2019bert} included to examine performance trade-offs in larger models.
We employed two main fine-tuning strategies: standard fine-tuning and PEFT methods, specifically Adapters, LoRA, and (IA)\(^3\), using the PEFT library \cite{peft-library}.

In non-DP settings, models were fined-tuned using the Hugging face trainer.
DP was implemented at a sample level using \texttt{dp-transformers}, a library built on Opacus and Hugging Face frameworks for DP fine-tuning LLMs \cite{sample-level, dp-transformers, huggingface, yousefpour2021opacus}.
In DP setups, PEFT methods were first integrated into the model, with DP mechanisms subsequently applied.
We evaluated models using privacy budgets ($\varepsilon$) of 1.0, 4.0, and 8.0 for DistilBERT and 4.0 for BERT-base, with a clipping norm of 1.5 and $\delta = 10^{-5}$ across all experiments.
Due to an incompatibility with DP-SGD, \textit{(IA)\(^3\) was excluded from DP experiments}: the (IA)\(^3\) implementation in the PEFT library did not produce the per-sample gradients required for DP training \cite{yousefpour2021opacus}.
For standard fine-tuning, DP-SGD was directly applied across all trainable parameters of the model.

Datasets used for this study included IMDb for sentiment analysis and QNLI for natural language inference.
Both datasets were tokenized and padded to a maximum sequence length of $256$ tokens using the Hugging Face library \cite{huggingface}.
Additionally, for efficient training on the larger QNLI dataset, we utilized a subset containing $50\%$ of the original training set, selected with a random seed of $42$.
Detailed training parameters, PEFT configurations, and DP settings are presented in tables \ref{tab:training_param} and \ref{table2: peft_config}, respectively.

\begin{table}[!h]
\centering

\begin{tabular}{|l|c|c|c|c|c|}
\hline
\textbf{Parameter}  & \textbf{DistilBERT} & \textbf{BERT-base} & \textbf{Adapter} & \textbf{LoRA} & \textbf{(IA)\(^3\)} \\ \hline
Batch Size          & 32                             & 32                            & 32              & 32           & 32           \\ \hline
Epochs              & 3                              & 3                             & 5               & 3            & 3            \\ \hline
Learning Rate (Non-DP) & 5e-5                        & 5e-5                          & 5e-4            & 5e-4         & 7e-3         \\ \hline
Learning Rate (DP)  & 7e-5                           & 7e-5                           & 1e-3            & 8e-4         & N/A          \\ \hline
Optimizer           & AdamW                          & AdamW                         & AdamW           & AdamW        & AdamW        \\ \hline
\end{tabular}
\caption{\footnotesize Training parameters for DistilBERT, BERT-base, and PEFT methods (Adapter, LoRA, and (IA)\(^3\)). This table outlines batch size, epochs, learning rates (for DP and non-DP settings), and optimizer configurations. Note that DP is not applicable for (IA)\(^3\) due to its incompatibility with per-sample gradient computations.}
\label{tab:training_param}
\end{table}

\begin{table}[!h]
\centering
\begin{tabular}{|l|l|}
\hline
\textbf{Method}     & \textbf{Configuration}                                                                 \\ \hline
Adapter             & Bottleneck size = 32                                                                   \\ \hline
LoRA                & Rank (r) = 8, LoRA alpha = 16, Dropout = 0.1, Target modules = \{‘q\_lin’, ‘v\_lin’\}  \\ \hline
(IA)\(^3\)                 & Target modules = \{‘q\_lin’, ‘v\_lin’, ‘out\_lin’\}, FF modules = \{‘out\_lin’\} \\ \hline
\end{tabular}
\caption{\footnotesize Special configurations for PEFT methods (Adapter, LoRA, and (IA)\(^3\)). This table specifies the bottleneck size for Adapter, rank (\(r\)), scaling factor, dropout, and target modules for LoRA, and the target and feedforward (FF) modules for (IA)\(^3\).}
\label{table2: peft_config}
\end{table}

\begin{figure}[!h]
    \centering
    \includegraphics[width=0.84\textwidth]{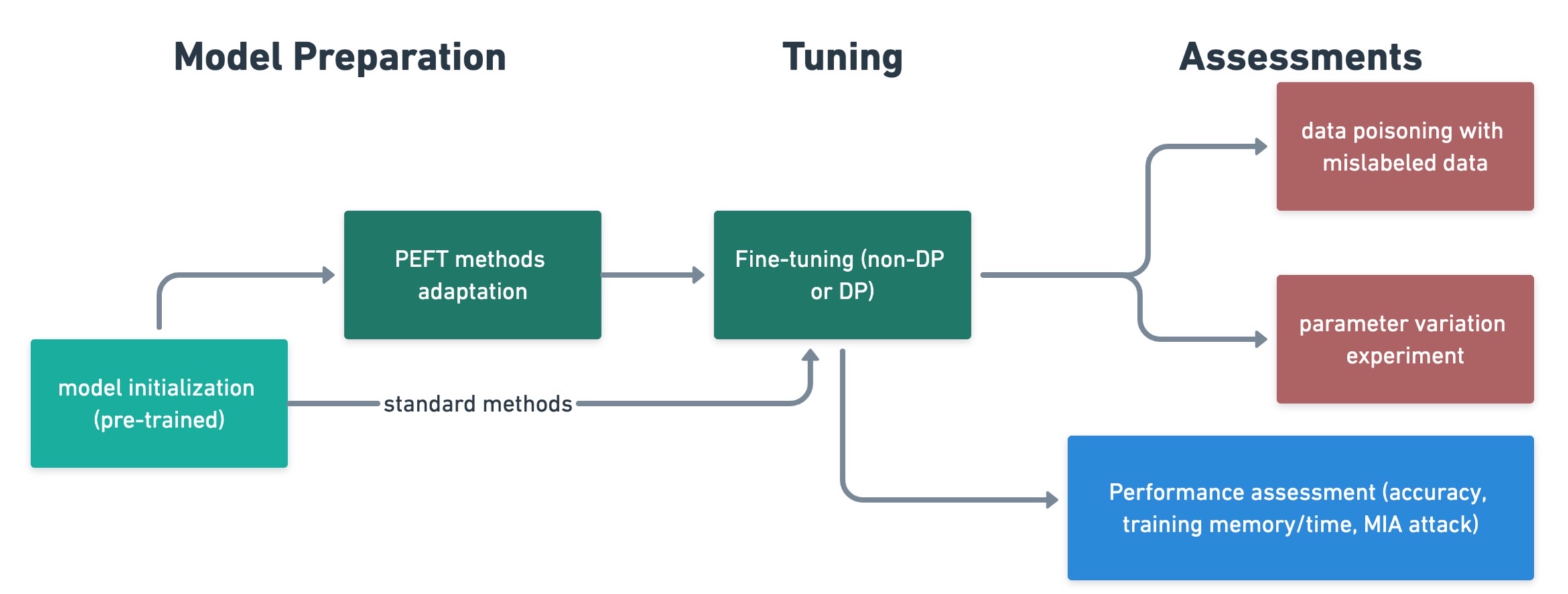}
    \caption{\footnotesize An overview of our method. Firstly, the pre-trained models are initialised, followed by the application of either PEFT methods or standard fine-tuning techniques. Secondly, the models are fine-tuned with/out DP. The final phase includes multiple evaluations: the poisoning attack to quantify memorisation, the parameter variation experiment to assess PEFT scalability, and performance assessment, covering accuracy, memory usage, training time, and vulnerability to MIAs.}
    \label{fig:process_illustration}
\end{figure}

\section{Results}

\subsection*{Performance Comparison: PEFT vs. Standard Fine-Tuning}

We evaluated standard fine-tuning and PEFT methods (LoRA, Adapter, (IA)\(^3\)) on IMDb and QNLI across non-DP and DP settings, focusing on task accuracy, memory usage, and training time per epoch.

\textbf{Accuracy:} As shown in Table \ref{tab:performance}, PEFT methods matched or exceeded the accuracy of standard fine-tuning.
In IMDb non-DP settings, LoRA closely aligned with DistilBERT’s accuracy while drastically reducing trainable parameters.
This trend held on QNLI, with PEFT methods achieving competitive accuracy levels.
Under DP constraints, LoRA even slightly outperformed DistilBERT and BERT-base.
At \(\varepsilon = 4.0\), LoRA achieved an accuracy of 85.3\%, compared to 84.3\% for DistilBERT and 81.1\% for BERT-base indicating robust performance in privacy-sensitive settings.

\textbf{Memory Efficiency and Training Speed:} PEFT methods provided substantial gains in memory and training efficiency.
As shown in Table \ref{tab:memory_training_speed}, BERT-base had the highest memory usage, whereas PEFT methods significantly reduced the memory overhead.
LoRA and Adapter used less than half the memory of BERT-base in both non-DP and DP settings.
Training speed followed a similar pattern, where PEFT methods completed epochs significantly faster than DistilBERT, even under DP training.

In summary, PEFT methods, particularly LoRA, offer an effective balance of task performance, memory, and computational efficiency.
Despite fewer parameters and a smaller base model, PEFT methods maintain accuracy comparable to standard fine-tuning while significantly reducing memory costs, making them well-suited for resource-constrained, privacy-sensitive applications.

\begin{table}[!h]
\centering
\begin{tabular}{|l|c|c|c|c|}
\hline
\textbf{Model}       & \textbf{Non-DP} & \textbf{DP (\(\varepsilon = 1.0\))} & \textbf{DP (\(\varepsilon = 4.0\))} & \textbf{DP (\(\varepsilon = 8.0\))} \\ \hline
\multicolumn{5}{|c|}{\textbf{IMDb}}                                                                                           \\ \hline
BERT-base            & \textbf{92.3}                & N/A                                   & 81.1                                     & N/A                                \\ \hline
DistilBERT           & 91.3                & 83.6                                 & 84.3                                 & 84.9                                 \\ \hline
LoRA                 & 90.7                & \textbf{84.7}                                 & \textbf{85.3}                                 & \textbf{85.9}                                 \\ \hline
Adapter              & 87.6                & 73.8                                 & 74.8                                 & 75.6                                 \\ \hline
(IA)\(^3\)            & 90.1                & N/A                                   & N/A                                   & N/A                                   \\ \hline
\multicolumn{5}{|c|}{\textbf{QNLI}}                                                                                           \\ \hline
BERT-base            & 88.5                & N/A                                 & 74.2                                 & N/A                                 \\ \hline
DistilBERT           & \textbf{86.2}                & N/A                                 & \textbf{76.4}                                & N/A                                 \\ \hline
LoRA                 & 85.2                & N/A                                 & 76.0                                 & N/A                                 \\ \hline
Adapter              & 82.0                & N/A                                 & 63.6                                 & N/A                                 \\ \hline
\end{tabular}
\caption{\footnotesize Performance comparison of PEFT methods and standard fine-tuning (accuracy in \%). This table shows the accuracy of PEFT methods (LoRA, Adapter, and (IA)\(^3\)) and standard fine-tuning for BERT-base and DistilBERT on IMDb and QNLI datasets under non-DP and DP settings (\(\varepsilon = 1.0, 4.0, 8.0\)). Results for (IA)\(^3\) and DP for BERT-base are marked as N/A due to technical constraints.}
\label{tab:performance}
\end{table}

\begin{table}[!h]
\centering
\begin{tabular}{|l|l|c|cc|cc|}
\hline
\multirow{2}{*}{\textbf{Dataset}} & \multirow{2}{*}{\textbf{Model}} & \multirow{2}{*}{\textbf{Trainable Parameters}} & \multicolumn{2}{c|}{\textbf{Memory (GB)}}    & \multicolumn{2}{c|}{\textbf{Training Speed (sec/epoch)}} \\ \cline{4-7} 
                                  &                                 &                                                   & \textbf{Non-DP}   & \textbf{DP}      & \textbf{Non-DP}         & \textbf{DP}          \\ \hline
\multirow{5}{*}{\textbf{IMDb}}    & BERT-base & 109,483,778  & 8.61  & 17.99 & 1067  & N/A                  \\ \cline{2-7} 
                                  & DistilBERT  & 66,955,010 & 4.49 & 8.50 & 753 & 1667                  \\ \cline{2-7} 
                                  & LoRA & 739,586 & 3.35 & \textbf{6.10}  & \textbf{652} & 698                  \\ \cline{2-7} 
                                  & Adapter  & 599,424   & \textbf{3.03} & 6.40 & 657 & \textbf{696}                  \\ \cline{2-7} 
                                  & (IA)\(^3\)& 605,954   & 3.42  & N/A  & 646 & N/A                  \\ \midrule

\multirow{4}{*}{\textbf{QNLI}}    & DistilBERT & 66,955,010   & 4.49 & 8.50 & 1086 & 1750                  \\ \cline{2-7} 
                                  & LoRA  & 739,586  & 3.40  & 5.95  & 947  & 1093                  \\ \cline{2-7} 
                                  & Adapter   & 599,424  & \textbf{3.02}   & \textbf{5.90}  & \textbf{945}  & \textbf{1006}                  \\ \cline{2-7} 
                                  & (IA)\(^3\)& 605,954   & 3.42  & N/A  & 946 & N/A                  \\ \hline

\end{tabular}
\caption{\footnotesize Comparison of trainable parameters, memory requirements, and training speed (GPU: P5000). This table presents the trainable parameters, memory usage (in GB), and training speed (in seconds per epoch) for different models and PEFT methods on IMDb and QNLI datasets under both non-DP and DP settings. Note that training speed for BERT-base under DP is not recorded as it utilized a larger GPU with higher memory capacity, making direct comparisons unsuitable.}
\label{tab:memory_training_speed}
\end{table}

\subsection*{Privacy Leakage: Membership Inference}

\begin{figure}[!h]
    \centering
    \begin{subfigure}[b]{0.41\textwidth}
        \centering
        \includegraphics[width=\textwidth]{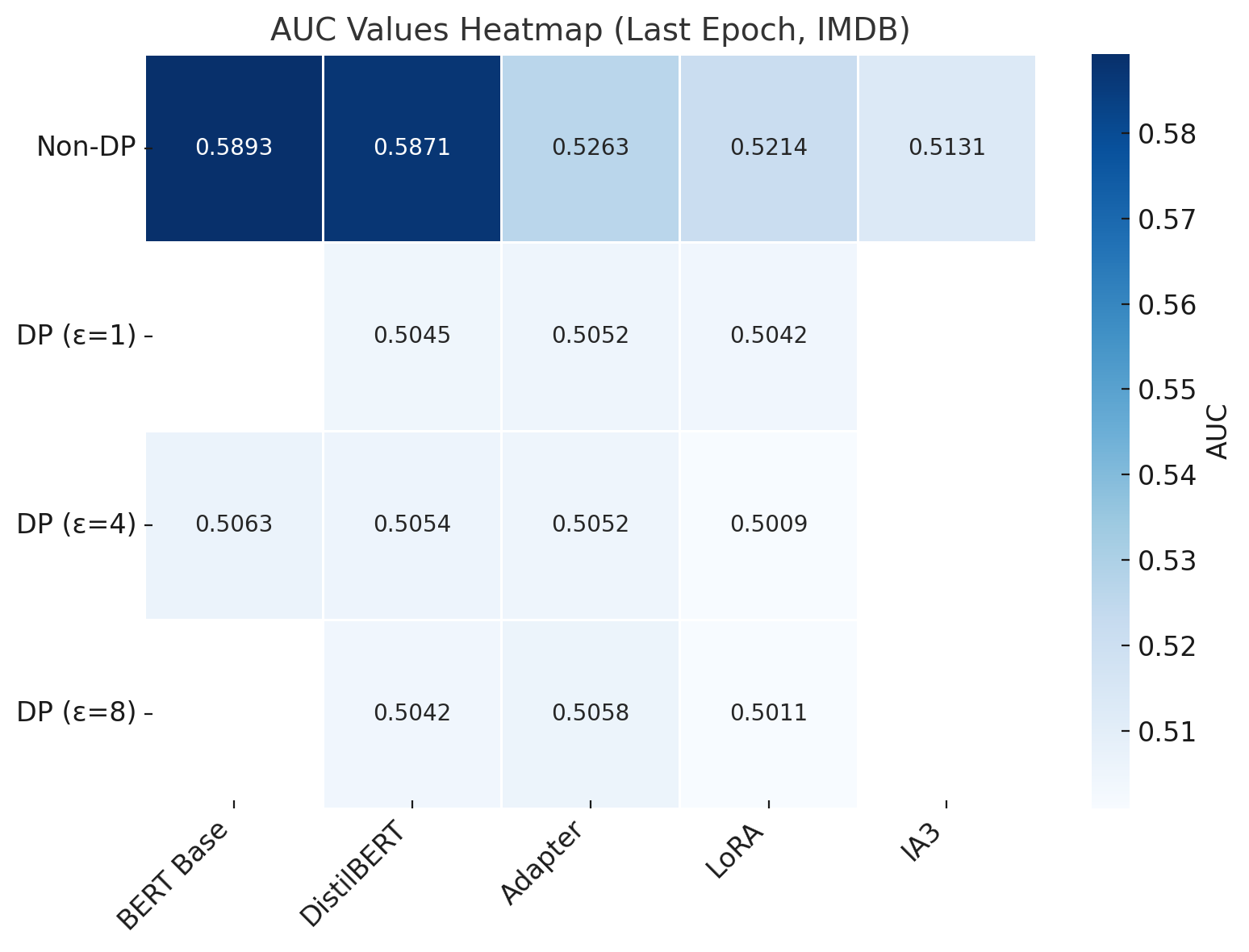}
        \label{fig:auc_imdb}
    \end{subfigure}
    \hfill
    \begin{subfigure}[b]{0.41\textwidth}
        \centering
        \includegraphics[width=\textwidth]{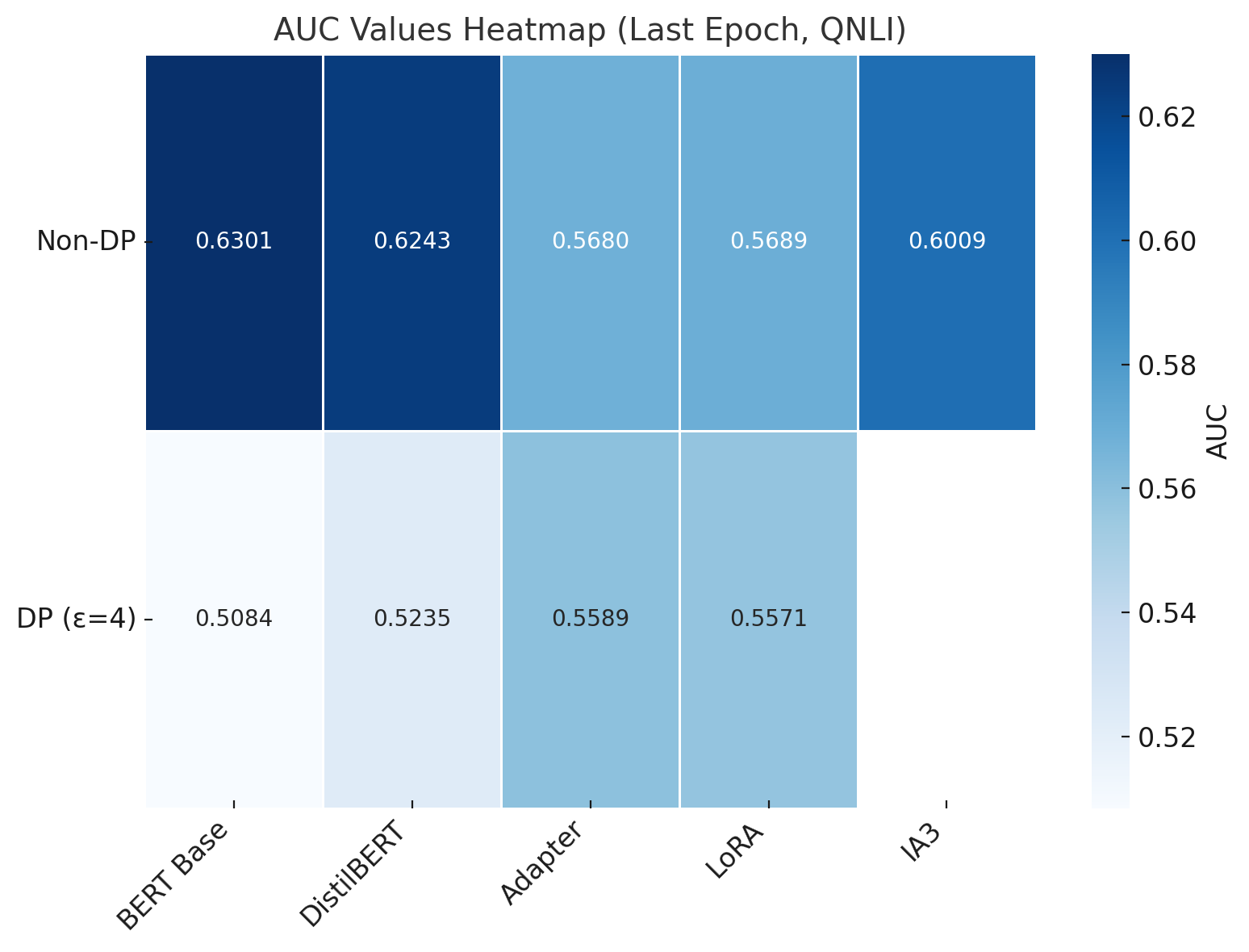}
        \label{fig:auc_qnli}
    \end{subfigure}
    \caption{\footnotesize AUC heatmaps for IMDb and QNLI datasets. These show the AUC values (last epoch) for different models and PEFT methods under non-DP and DP settings (\(\varepsilon = 1.0, 4.0, 8.0\)). The left heatmap corresponds to the IMDb dataset, while the right heatmap corresponds to QNLI. Darker shades represent higher AUC values, indicating greater privacy leakage and lighter shades represent lower AUC values, suggesting reduced memorisation.}
    \label{fig:auc_heatmaps}
\end{figure}

Using a loss-based black-box MIA, we quantified the risk of privacy leakage by calculating the AUC score at the final training epoch.
As illustrated in Figure \ref{fig:auc_heatmaps}, standard fine-tuning methods demonstrated greater susceptibility to MIAs compared to PEFT methods.
Specifically, non-DP DistilBERT reaching an AUC of $0.59$ on the IMDb dataset, indicating increased memorisation and privacy risks.
In contrast, PEFT methods like LoRA and Adapter showed lower AUC scores of $0.52$ and $0.53$, respectively, underscoring their enhanced resilience against membership inference attacks.

A notable trend was observed with respect to the model size (i.e. their capacity): larger models, consistently exhibited higher AUC values and higher privacy leakage compared to DistilBERT, particularly in the non-DP settings, suggesting that larger models may inherently be more prone to memorisation, amplifying privacy risks in the absence of DP.

Under DP constraints, privacy leakage diminished across all settings, with DistilBERT’s AUC on IMDb dropping down to $0.50$ at \(\varepsilon = 8.0\), demonstrating DP’s efficacy in reducing memorisation.
However, PEFT methods showed smaller AUC reductions under DP, implying a weaker interaction with DP mechanisms (in contrast to standard fine-tuning, where the reduction is significant).
This pattern was especially evident on QNLI, where LoRA and Adapter maintained relatively stable AUC values, slightly higher than their standard fine-tuning counterpart. 
The reduced parameter updates in PEFT methods concentrate DP noise on a smaller subset of parameters, which may reduce the overall effectiveness of DP in mitigating privacy leakage compared to standard fine-tuning, where noise is distributed across all parameters.
This observation suggests the need to develop specialised DP strategies that account for PEFT’s unique structure, ensuring that noise is applied more effectively to enhance privacy protection while maintaining computational efficiency.

\subsection*{Privacy Leakage: Auditing using canaries}

Using the mislabelled samples (i.e. our canaries), we examined how the models responded to a controlled set of poisonous data points, focusing on whether they had memorised the incorrect feature-label associations.

\begin{figure}[!h]
    \centering
    \begin{subfigure}[b]{0.43\textwidth}
        \centering
        \includegraphics[width=\textwidth]{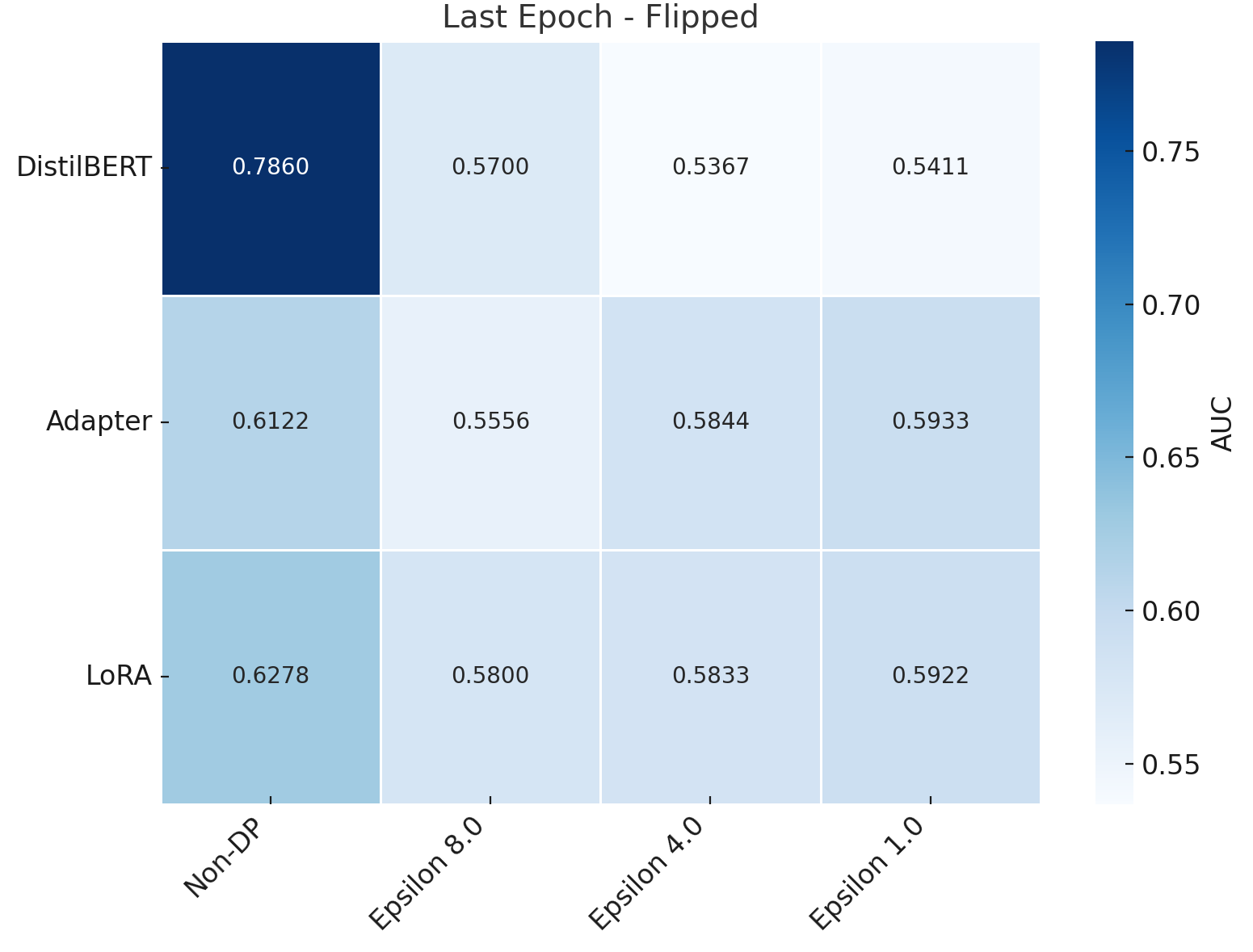}
        \label{fig:memo_imdb}
    \end{subfigure}
    \begin{subfigure}[b]{0.43\textwidth}
        \centering
        \includegraphics[width=\textwidth]{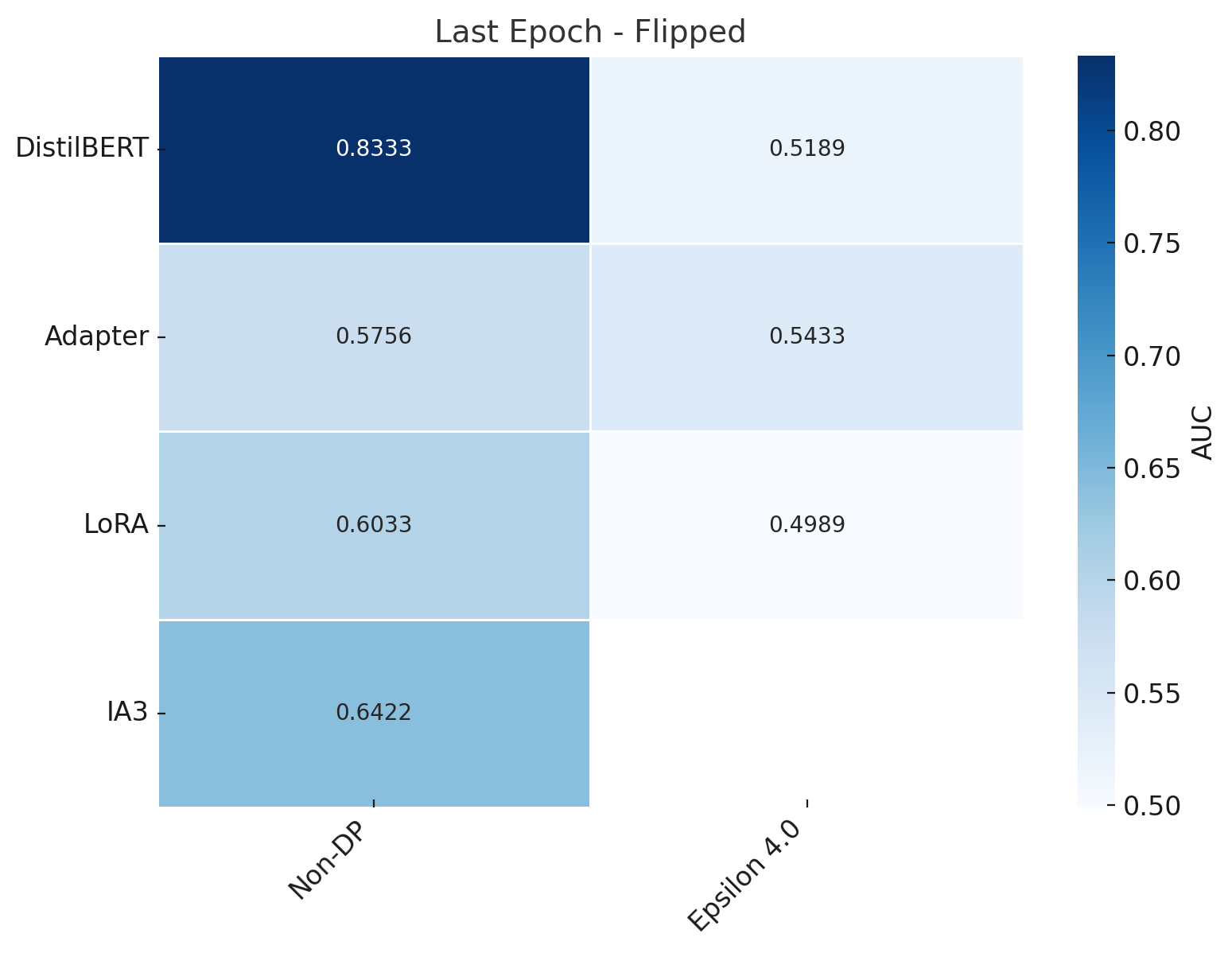}
        \label{fig:memo_qnli}
    \end{subfigure}
    \caption{\footnotesize Memorisation of IMDB and QNLI datasets. These heatmaps show AUC values (last epoch) for different models and PEFT methods under non-DP and DP settings on IMDb (left) and QNLI (right) datasets. Data poisoning experiments were used to assess memorisation risks, where higher AUC values indicate greater memorisation of mislabelled data. Darker shades represent higher memorisation.}
    \label{fig:memorisation_heatmaps}
\end{figure}

As shown in Figure \ref{fig:memorisation_heatmaps}, standard fine-tuning with DistilBERT on the IMDb dataset exhibited a high degree of memorisation in non-DP settings, with elevated AUC values on the flipped subset, indicating a greater risk of privacy leakage.
In contrast, PEFT methods like LoRA and Adapter maintained significantly lower AUC scores, demonstrating reduced memorisation and a stronger resilience against privacy attackers.
This suggests that, in non-DP settings, PEFT methods effectively limit the sensitivity (in a non-DP sense) of the model to outliers, enhancing privacy preservation.
A similar trend was observed on QNLI, where non-DP DistilBERT also showed higher degree of memorisation, while PEFT methods mitigated this effect, reinforcing the privacy advantages of PEFT.

In DP settings, the AUC scores on the poisoned subset decreased across all methods, indicating DP’s effectiveness in reducing memorisation.
However, similar to the previous experiments, PEFT methods like LoRA and Adapter showed less reduction in AUC under DP compared to standard fine-tuning, suggesting that DP mechanisms are less effective in mitigating memorisation for PEFT methods.
This is likely the case since DP-SGD can often act as a \say{natural regulariser}, limiting the impact that PEFT methods (partially adressing the same issue) may have.
Additionally, PEFT updates only a subset of parameters, limiting the noise addition to a smaller parameter set.
This means that while the new information obtained during fine-tuning may be memorised less, the initial pre-training data can still be inferred and should the features (or even datapoints) in these datasets overlap, DP training would not be able to offer \textbf{meaningful} protection against MIAs (while still providing meaningful guarantees with respect to fine-tuning information bounds).

\begin{figure}[!h]
    \centering
    \begin{subfigure}[b]{0.45\textwidth}
        \centering
        \includegraphics[width=\textwidth]{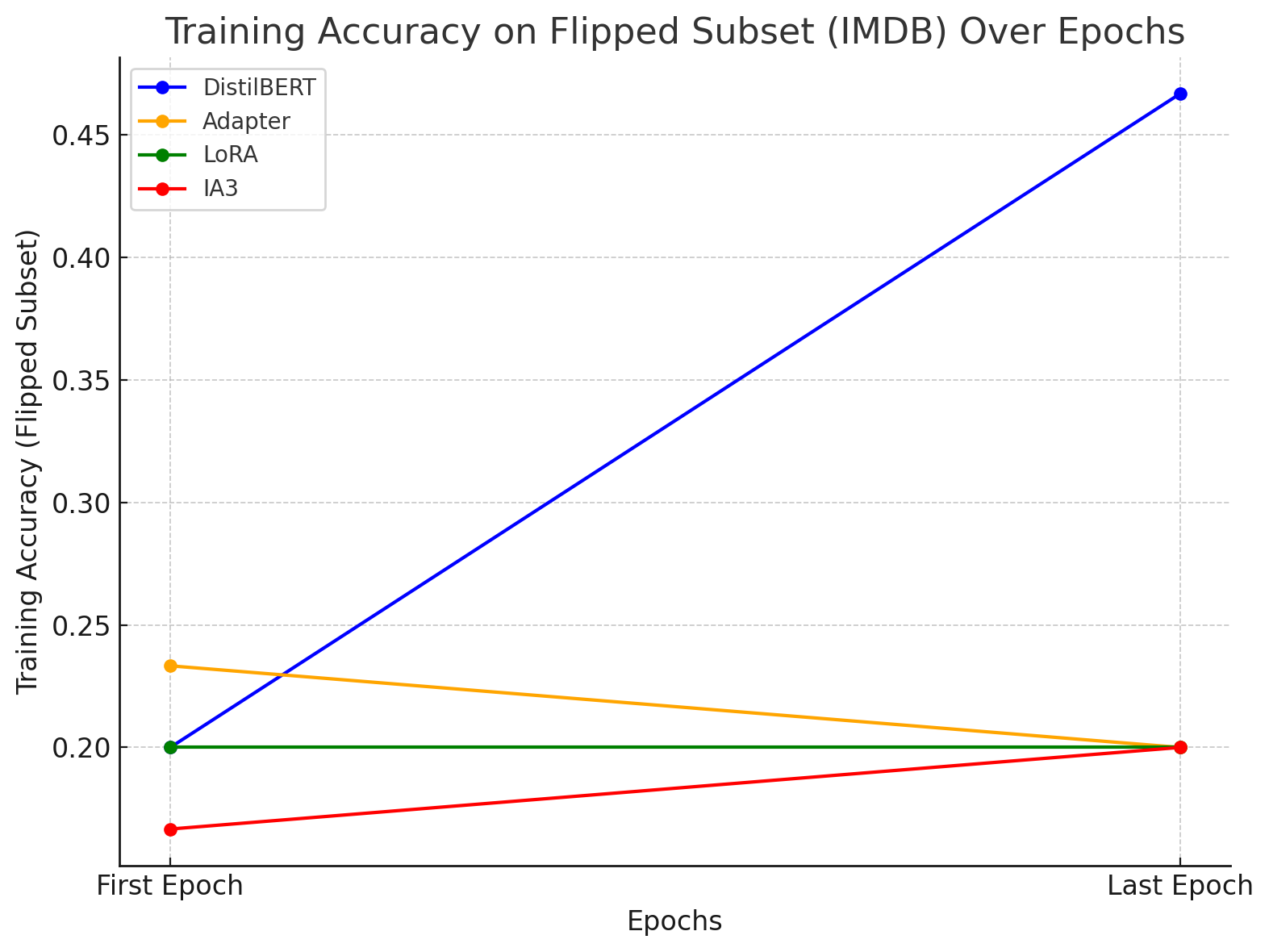}
        \label{fig:memo_acc_imdb}
    \end{subfigure}
    \begin{subfigure}[b]{0.45\textwidth}
        \centering
        \includegraphics[width=\textwidth]{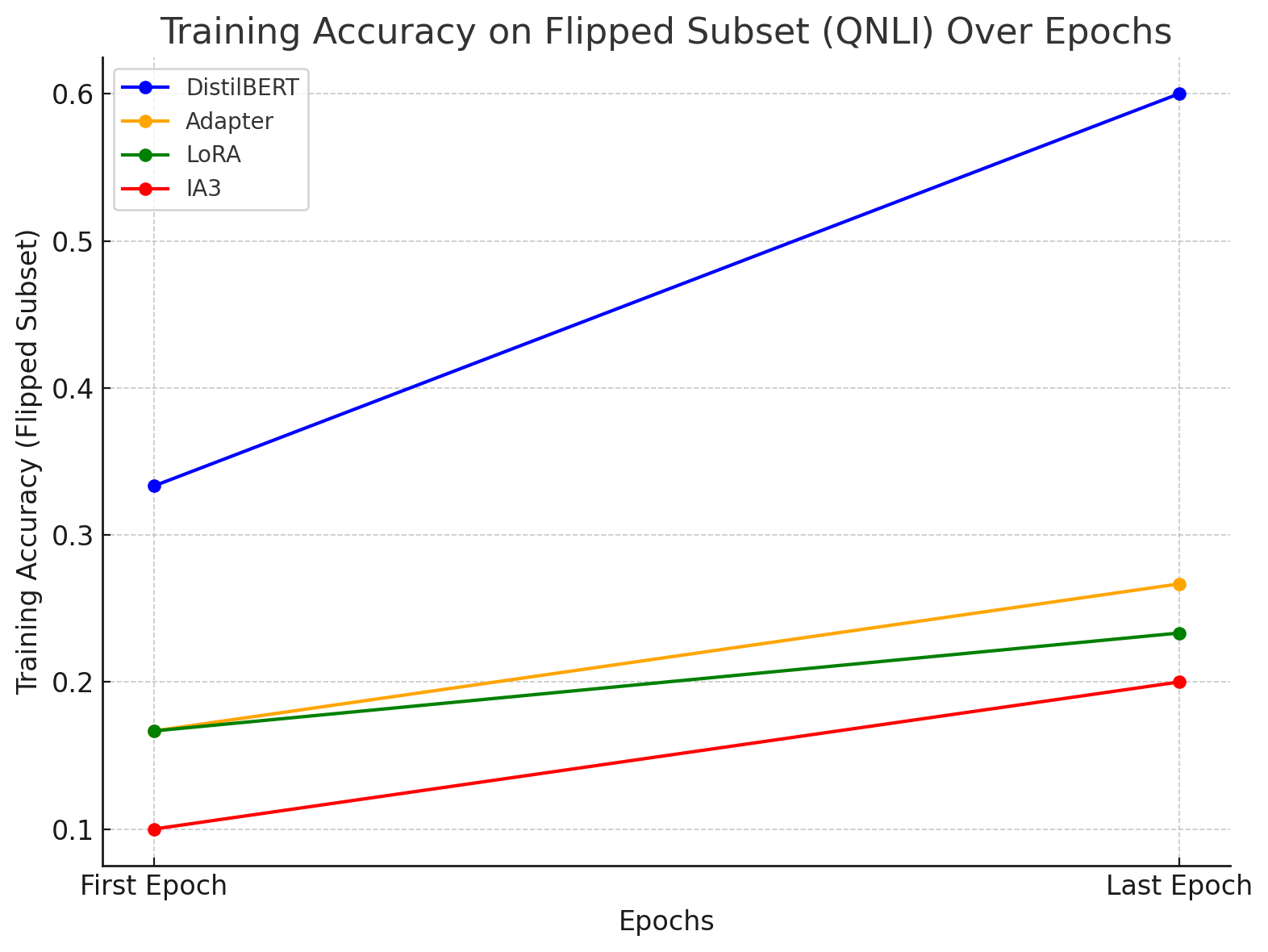}
        \label{fig:memo_acc_qnli}
    \end{subfigure}
    \caption{\footnotesize Training accuracy on the flipped subsets for IMDb and QNLI. This figure plots the training accuracy over epochs on poisoned subsets of the IMDb (left) and QNLI (right) datasets, comparing DistilBERT, Adapter, LoRA, and (IA)\(^3\), highlights the model’s tendency to memorise mislabeled data.}
    \label{fig:memorisation_accuracy}
\end{figure}

We also tracked training accuracy on the flipped subset across epochs for both IMDb and QNLI datasets, as shown in Figure \ref{fig:memorisation_accuracy}.
Standard fine-tuning with DistilBERT revealed a marked increase in accuracy on the poisoned subset, indicating a higher memorisation capacity on this data. 
For instance, on IMDb, DistilBERT’s accuracy rose notably from $0.20$ in the first epoch to over $0.45$ by the final epoch, suggesting greater sensitivity to outliers.
In contrast, PEFT methods such as LoRA, Adapter, and (IA)\(^3\) maintained consistently low accuracy on the mislabelled subset, with only minor increases over time.
This stability reflects their higher robustness against memorisation, as they are less prone to retaining incorrect label associations.
For example, LoRA’s accuracy on IMDb remained consistently low, and on QNLI, PEFT methods stayed below $0.40$, whereas DistilBERT’s accuracy climbed sharply to $0.60$, underscoring PEFT’s improved generalisation and resistance to memorising outliers.

\subsubsection*{PEFT Parameter Variation}

\begin{figure}[!h]
    \centering
    \begin{subfigure}[b]{0.45\textwidth}
        \centering
        \includegraphics[width=\textwidth]{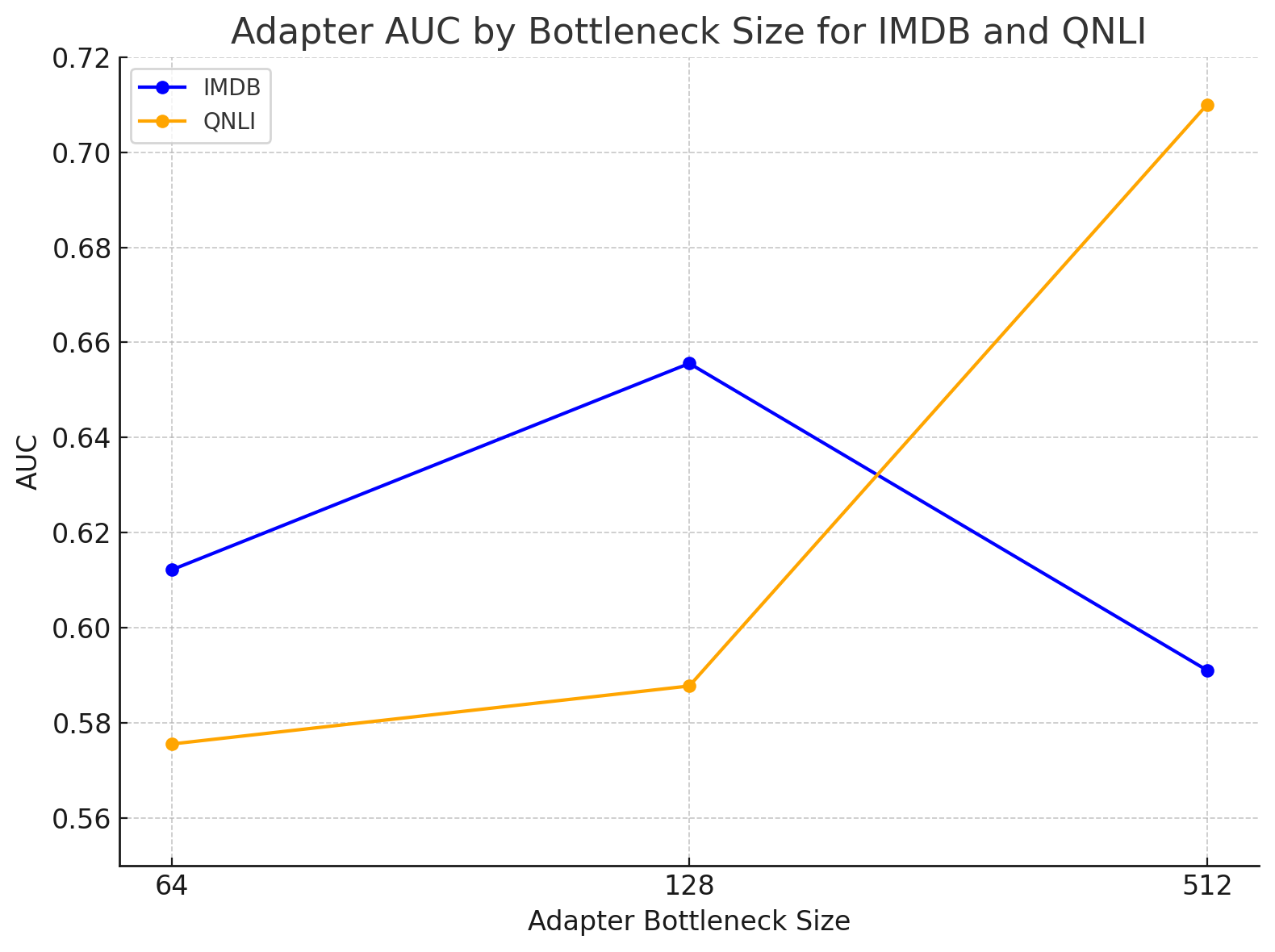}
        \label{fig:param_inc_imdb}
    \end{subfigure}
    \begin{subfigure}[b]{0.45\textwidth}
        \centering
        \includegraphics[width=\textwidth]{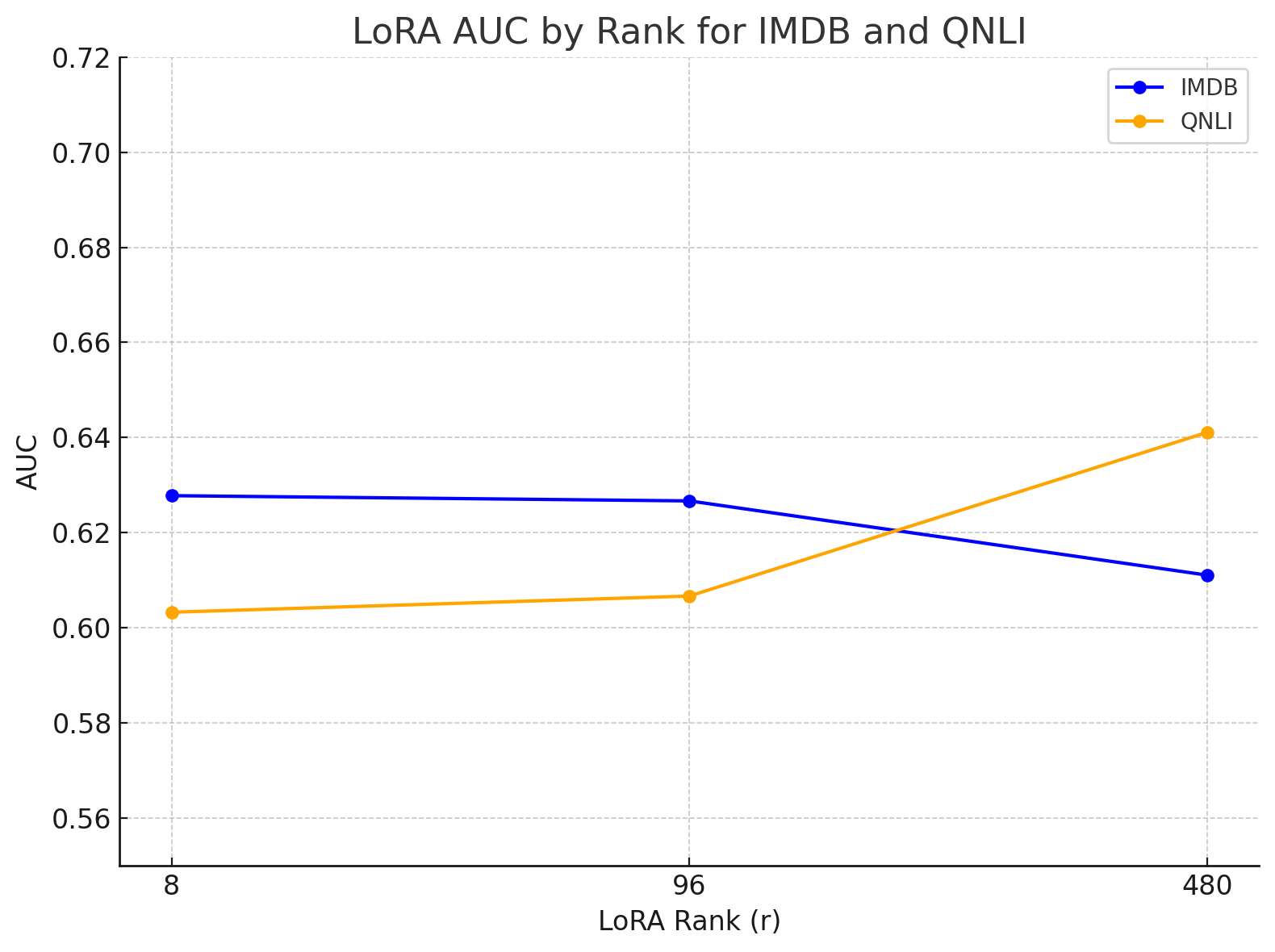}
        \label{fig:param_inc_qnli}
    \end{subfigure}
    \caption{\footnotesize AUC results with increasing parameter count on IMDB and QNLI. The Adapter plot (left) shows AUC scores with varying bottleneck sizes, which control the intermediate feature size within Adapter layers, inserted into FF layers. The LoRA plot (right) shows AUC scores across different rank \textit{r} values, which define the low-rank matrices dimensionalities added to attention layers.}
    \label{fig:param_auc}
\end{figure}

We tested the hypothesis that PEFT methods are more robust against memorisation only due to their reduced parameter count by varying the parameter count in Adapter and LoRA models.
As shown in Figure \ref{fig:param_auc}, increasing Adapter’s bottleneck size led to higher AUC scores on QNLI, rising from $0.58$ to $0.71$.
However, this pattern was inconsistent on IMDb, where AUC scores fluctuated, suggesting that Adapter’s sensitivity to the number of trainable parameters may vary based on the dataset. 
For LoRA, increasing the \textit{r} values, which controls dimensionality of introduced low-rank matrices from $8$ to $480$ showed minimal AUC variation across both datasets, reinforcing its robustness against memorisation.
This stability contrasts with Adapter, highlighting LoRA’s capacity to reduce memorisation risks even with more parameters.

Overall, while PEFT methods, and especially LoRA, exhibit strong resistance to MIA attacks and lower associated memorisation risks compared to standard fine-tuning, we found that this robustness does not solely depend on the umber of training parameters.

\section{Discussion}

Our findings underscore the effectiveness of PEFT methods, particularly LoRA, as practical alternatives to standard fine-tuning in privacy-sensitive settings.
These methods not only maintain high task-specific accuracy, but also dramatically reduce parameter count and memory usage, making it feasible to deploy privacy-preserving models even in resource-limited environments.
This efficiency directly addresses critical limitations of standard fine-tuning, supporting robust model performance in scenarios where computational resources might otherwise limit adoption of LLMs and private fine-tuning of such models.

\textbf{Privacy Resilience Without DP:} In non-DP settings, PEFT methods showed reduced privacy leakage, evidenced by lower MIA AUC scores.
This suggests that their parameter-efficient design limits the model's capacity to memorise individual training data points, inherently reducing empirical memorisation risks even without formal privacy protection mechanisms. 
This observation aligns with recent findings showing that LoRA effectively preserves pre-trained knowledge, which in turn reduces the risk of memorising task-specific fine-tuning data \cite{Biderman2024LoRA}.
The fact that PEFT methods can mitigate privacy leakage without DP could make them particularly appealing in real-world applications where DP deployment may be costly or impractical due to constrained computational budget.
Our results position PEFT as both a stand-alone privacy-preserving measure and a complementary solution in cases where full-model fine-tuning with DP might be infeasible.
By reducing memorisation without requiring DP, PEFT methods broaden the options for model deployment in real-world applications with fewer privacy risks stemming from fine-tuning memorisation.

\textbf{Investigating Induced Memorisation:} The data poisoning experiment experiment provided compelling evidence of PEFT’s lower tendency to memorise outlying, often more sensitive data.
Lower AUC and accuracy scores on the flipped subsets for LoRA and Adapter, as compared to standard fine-tuning, indicate a reduced risk of memorisation, enhancing privacy protection for unique or rare data points.The results from this experiment show that PEFT methods can potentially limit model memorisation of unintended patterns, suggesting a greater resilience against privacy risks even without additional privacy controls.

\textbf{DP's Weaker Effectiveness on PEFT Methods:}  While DP effectively reduced memorisation in standard fine-tuning, its impact on PEFT methods was comparatively weaker.
Specifically, PEFT models retained higher AUC values in DP settings compared to standard fine-tuning, indicating that DP mechanisms may not interact as effectively with PEFT’s parameter-efficient structure.
This may be due to PEFT’s focus on updating only a small subset of parameters, which can concentrate the DP noise within these limited updates, and potentially reduce DP’s ability to mitigate privacy risks across the entire model.
These findings highlight the need to develop DP strategies optimised for PEFT architectures, potentially by dynamically adjusting noise distribution to enhance privacy protection across the full model.
Such approaches could maximise privacy utility in sensitive applications while maintaining computational efficiency.

\textbf{Parameter Variation and Robustness to Memorisation:} Our parameter variation experiments revealed that PEFT robustness may not solely come from a reduced parameter count.
LoRA and Adapter maintained low AUC scores across configurations with increased parameters, demonstrating limited increases in privacy leakage even with more parameters. 
These findings highlight that other factors, such as how and where PEFT modules are integrated within the model, could play a significant role in privacy preservation (i.e. adaptations of the model similar to \cite{usynin2022zen}).
For example, embedding PEFT modules within attention mechanisms (LoRA) versus FF layers (Adapters), or within different attention subcomponents (key-value pairs, query vectors or output layers) \cite{hu2021lora, houlsby2019adapter}.
Exploring these configurations could yield deeper insights into effective privacy-preserving strategies for fine-tuning LLMs.

\textbf{Limitations and Future Directions:} Our study focused on specific models (BERTs), PEFT methods (LoRA, Adapter, (IA)\(^3\)) and datasets (IMDb, QNLI), which may limit the generalizability of our conclusions. Future work should expand the scope by exploring a wider variety of PEFT configurations and task domains to validate the consistency of these privacy-preserving benefits.
Additionally, adaptive DP mechanisms tailored specifically to PEFT structures could improve privacy without sacrificing model utility, laying the groundwork for more secure and efficient fine-tuning practices in increasingly privacy-sensitive applications.

Another potential direction for future research involves examining the robustness of PEFT methods when applied to biased datasets or minority groups.
Assessing PEFT's privacy and utility on imbalanced or skewed data distributions will be essential for understanding its effectiveness in protecting underrepresented groups from privacy risks \cite{truex2019effects}.
In addition, expanding the range of privacy assessment techniques to include broader methods, such as the model inversion attacks, which aim to reconstruct sensitive training data from intermediate representations of the model, could provide a more comprehensive evaluation of privacy risks associated with PEFT methods under DP \cite{haim2022reconstructing}. 
Furthermore, incorporating advanced inference techniques like RMIA \cite{zarifzadeh2023low} would allow us to test PEFT methods against stronger membership inference scenarios, allowing us to consider different threat models where the adversary is able to satisfy the assumption of having access to a certain (small) number of real data points, significantly reducing their computation requirements stemming from the need to train many shadow models.
As the demand for privacy-preserving LLMs grows, these advancements will support more ethical, secure, and resource-efficient deployments in sensitive applications.

\section{Conclusions}

This study highlights the potential of EFT methods, particularly LoRA and Adapter, as both complementary and alternative approaches to standard fine-tuning for privacy-sensitive applications.
Our results show that PEFT methods preserve task-specific performance while reducing memory usage and privacy leakage.
The data poisoning experiment provided direct evidence of PEFT’s reduced memorisation tendencies, reinforcing their suitability for privacy-sensitive scenarios.
While combining PEFT with DP may provide optimal privacy protection, PEFT alone offers valuable privacy benefits when DP training is infeasible.

Furthermore, our findings reveal that PEFT’s robustness against privacy risks may not solely depend on fewer parameters but could also be influenced by its architectural design.
Future research should explore the impact of PEFT’s architectural placement within large language models, particularly within attention and FF layers, to further optimize their privacy-preserving capabilities.
In DP settings, while PEFT methods generally reduced privacy leakage, their interaction with DP was less pronounced than with the standard fine-tuning, suggesting potential avenues for refining DP-PEFT compatibility. These insights pave the way for more resource-efficient, secure fine-tuning practices in privacy-critical applications.

\bibliographystyle{unsrt}  
\bibliography{references}  
\appendix

\end{document}